\documentclass[letterpaper, 10 pt, conference]{ieeeconf}  

\IEEEoverridecommandlockouts                              

\makeatletter
\let\NAT@parse\undefined
\makeatother

\title{\LARGE \bf
Reducing Oracle Feedback with Vision-Language Embeddings for Preference-Based RL
}

\author{Udita Ghosh$^{1*}$, Dripta S. Raychaudhuri$^{2}$, Jiachen Li$^{1}$, Konstantinos Karydis$^{1}$, Amit Roy-Chowdhury$^{1}$
\author{$^{1}$Univeristy of California, Riverside, $^{2}$AWS AI Labs}
\thanks{$^{1}$Univeristy of California, Riverside; $^{2}$AWS AI Labs (Work done outside AWS)}%
\thanks{* Corresponding author: ughos002@ucr.edu}
}
\usepackage{amsmath}
\usepackage{microtype}
\usepackage{amsfonts}
\usepackage{caption}
\usepackage{graphicx}
\usepackage{amsmath}
\usepackage{makecell}

\usepackage[pagebackref,breaklinks,colorlinks]{hyperref}

\usepackage{xcolor}

\newcommand{\algo}{\textrm{ROVED}}

\begin{document}

\maketitle
\thispagestyle{empty}
\pagestyle{empty}

\begin{abstract}

Preference-based reinforcement learning can learn effective reward functions from comparisons, but its scalability is constrained by the high cost of oracle feedback. Lightweight vision-language embedding (VLE) models provide a cheaper alternative, but their noisy outputs limit their effectiveness as standalone reward generators. To address this challenge, we propose \algo, a hybrid framework that combines VLE-based supervision with targeted oracle feedback. Our method uses the VLE to generate segment-level preferences and defers to an oracle only for samples with high uncertainty, identified through a filtering mechanism. In addition, we introduce a parameter-efficient fine-tuning method that adapts the VLE with the obtained oracle feedback in order to improve the model over time in a synergistic fashion. This ensures the retention of the scalability of embeddings and the accuracy of oracles, while avoiding their inefficiencies. Across multiple robotic manipulation tasks, \algo~matches or surpasses prior preference-based methods while reducing oracle queries by up to 80\%. Remarkably, the adapted VLE generalizes across tasks, yielding cumulative annotation savings of up to 90\%, highlighting the practicality of combining scalable embeddings with precise oracle supervision for preference-based RL. Project page:~\url{https://roved-icra-2026.github.io/}

\end{abstract}

\section{Introduction} \label{sec:intro}

When a robot manipulator is assigned a task, it must possess the necessary skills to complete it effectively. Acquiring such skills through reinforcement learning (RL) is challenging, primarily because designing a reward function that reliably guides the agent toward the desired behavior is difficult. Hand-crafted rewards often require significant domain expertise, risk unintended shortcuts or reward hacking, and struggle to capture complex task goals.
A widely adopted alternative is preference-based RL (PbRL)~\cite{christiano2017deep,lee2021pebble}, where a reward function is inferred from human or oracle feedback in the form of comparisons between agent behaviors. This approach removes the need for carefully engineered reward functions and provides a more direct way to align agent behavior with human intent. However, current PbRL methods typically demand large volumes of high-quality feedback, which is costly and time-consuming to obtain, thereby limiting their practicality and scalability.


To reduce human effort, recent work has explored vision–language embedding (VLE) models such as CLIP~\cite{radford2021learning}, which enable zero-shot reward estimation from natural language descriptions (\textit{e.g.}, ``open the door'') by mapping text and images into a shared representation space~\cite{rocamonde2023vision,ma2023liv,sontakke2024roboclip,lidecisionnce}. While scalable, these VLE-based rewards are often coarse and noisy, limiting their utility in precise domains such as robotic manipulation~\cite{fu2024furl}. More recently, large vision–language models (VLM) like Gemini-Pro~\cite{team2024gemini} have been proposed as automated oracles for preference learning~\cite{wang2024}, replacing human annotators entirely. Although accurate, VLMs incur high query costs and suffer from slow, auto-regressive inference. This motivates the need for a \emph{method that retains the scalability of embedding based approaches and the accuracy of oracles, while avoiding their inefficiencies.}

Towards this goal, we introduce \algo: \textbf{R}educing \textbf{O}racle Feedback using \textbf{V}ision-language \textbf{E}mbed\textbf{D}ings, a framework for efficient PbRL that combines lightweight VLE models with targeted oracle feedback (where the oracle may be a human or a VLM). Prior approaches rely exclusively on either embeddings~\cite{rocamonde2023vision} or oracle comparisons~\cite{christiano2017deep,wang2024}. In contrast, \algo~adopts a hybrid strategy: the VLE generates segment-level preferences and defers to the oracle only when confidence is low. This creates a feedback loop where the VLE improves from oracle supervision and becomes increasingly selective in querying, thereby preserving annotation quality while substantially reducing query costs.

To the best of our knowledge, we are the first to exploit VLEs as a supplementary source of noisy labels to reduce dependence on high-quality oracle feedback. Our approach rests on two components. \textit{First}, we improve the quality of scalable preference labels through a parameter-efficient fine-tuning scheme that combines an unsupervised dynamics-aware objective with sparse oracle feedback. Prior efforts have attempted to denoise embedding-based rewards using expert demonstrations or environment shaping~\cite{fu2024furl}, but demonstrations are costly to collect~\cite{akgun2012keyframe} and often suffer from domain gaps~\cite{smith2019avid}. In contrast, preference feedback is lightweight, accessible, and more broadly applicable~\cite{hejna2023few}. \textit{Second}, to minimize oracle queries, we adopt robust confidence-aware training~\cite{song2022learning,cheng2024rime}: by constraining the KL divergence between the reward model and VLE predictions, the system automatically resolves confident cases and escalates only uncertain ones. Together, these mechanisms deliver both efficiency and precision in preference-based reward learning.

We evaluate \algo~on robotic manipulation tasks from Meta-World~\cite{yu2020meta} and demonstrate that it learns reward functions capable of training effective policies. \algo~matches the performance of oracle-only methods while reducing annotation cost by \textbf{50-80\%}. Moreover, a VLE fine-tuned on one task generalizes to related tasks with minimal additional supervision, yielding \emph{cumulative annotation savings} of \textbf{75-90\%}. This cross-task generalization—absent in prior work—demonstrates that pairing scalable VLE models with selective oracle feedback is a promising direction for practical and efficient PbRL. The main contributions of this work are:

\begin{itemize}
\item We present \algo, a novel framework for efficient PbRL that combines the scalability of VLEs with the precision of selective oracle supervision. 
\item We introduce two key techniques: (i) a parameter-efficient adaptation scheme that improves noisy VLE preference labels with dynamics-aware objectives and sparse oracle feedback, and (ii) an uncertainty-aware sample selection strategy that queries the oracle only for uncertain cases in an effort to reduce oracle feedback.
\item We demonstrate that \algo~achieves oracle-level performance on Meta-World tasks while cutting annotation cost by 50–80\%, and further achieves 75–90\% cumulative savings through cross-task generalization.
\end{itemize}

\section{Related Work} \label{sec:related_works}

\noindent \textbf{Designing reward functions.} 
Traditional methods for reward design often rely on manual trial-and-error, requiring substantial domain expertise and struggling to scale to complex, long-horizon tasks. Inverse reinforcement learning offers an alternative by attempting to infer reward functions from expert demonstrations~\cite{ho2016generative}. Evolutionary algorithms have also been explored for automated reward function discovery~\cite{chiang2019learning}. More recently, foundation models have enabled reward design from high-level task descriptions. Large language models (LLM) have been employed to derive reward functions directly from natural language task descriptions~\cite{ma2023eureka,wang2025prefclm}, although these approaches assume access to the environment's source code. VLE models provide another path, aligning visual observations with textual descriptions to yield dense reward signals~\cite{rocamonde2023vision,ma2023liv,sontakke2024roboclip,lidecisionnce}. These methods are fast and scalable, but VLE rewards are typically noisy and insufficiently precise for fine-grained control~\cite{fu2024furl}.

\noindent \textbf{Preference-based RL.} 
Preference feedback has been widely explored across domains, including natural language processing~\cite{rafailov2024direct} and robotics~\cite{lee2021pebble}. In RL, Christiano \textit{et al.} pioneered the use of human trajectory comparisons~\cite{christiano2017deep}, with subsequent works improving sample efficiency and generalization. For instance, PEBBLE~\cite{lee2021pebble} combined human preferences with unsupervised exploration, SURF~\cite{park2022surf} augmented preference datasets using semi-supervised learning, and MRN~\cite{liu2022meta} incorporated policy performance into reward model updates. These methods leverage the fact that relative judgments are easier for humans than explicit reward specification, but they remain constrained by costly human annotation. To alleviate this bottleneck, recent work such as RL-VLM-F~\cite{wang2024} uses large VLMs, such as Gemini-Pro, as preference oracles for preference learning. While promising in accuracy, this introduces new challenges: VLMs are expensive to query due to API costs and slow because of their auto-regressive nature.

Given these scalability challenges, we propose a hybrid framework that selectively queries expensive oracles only when fast, automatic supervision is unreliable. Specifically, we use a lightweight VLE model to generate initial preference labels, deferring to the oracle based on a model uncertainty criterion. Unlike prior methods that fully commit to either oracle or automated feedback, our approach dynamically balances scalability and precision by allocating supervision where it is most needed, following principles from uncertainty-based active learning~\cite{li2024survey}. This targeted strategy substantially reduces annotation costs without sacrificing reward quality.

\noindent \textbf{Learning from noisy labels.} 
The challenge of learning from noisy or imprecise labels has been extensively studied in supervised learning, particularly with the rise of machine-generated annotations~\cite{wang2022self}. Robust training methods tackle this issue through diverse strategies, including regularization techniques~\cite{lukasik2020does, ghosh2025robust}, specialized loss functions~\cite{zhang2018generalized}, and sample-selection mechanisms~\cite{wang2021denoising}. In our framework, the preference labels generated by the VLE model may be noisy due to its inherent limitations. To mitigate this, we adopt a sample-selection strategy inspired by RIME~\cite{cheng2024rime}, which identifies confident labels for training while flagging uncertain samples for oracle refinement. This combination of robust training and targeted oracle feedback ensures that our reward model remains accurate while keeping preference learning efficient.

\begin{figure*}[!t]
    \centering
    \includegraphics[width=0.95\textwidth]{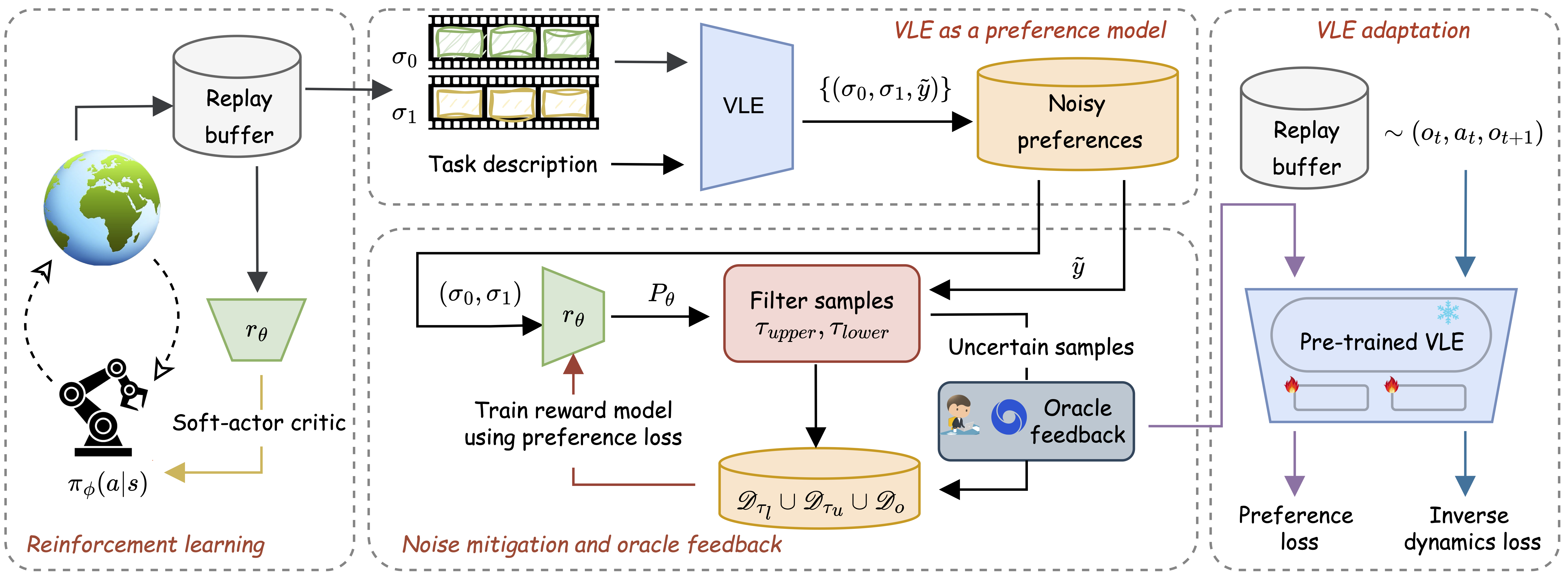}
    \caption{\textbf{Overview of our approach.} Given a task description, \algo~iteratively updates the policy $\pi_\phi$ via reinforcement learning using the reward model $r_\theta$. Trajectory segments from the replay buffer are sampled and labeled with VLE-generated preferences. These samples are then classified as clean or noisy using thresholds $\tau_{upper}$ and $\tau_{lower}$. A budgeted subset of noisy samples is sent for oracle annotation. The reward model is trained on both VLE and oracle-labeled preferences, while the VLE is fine-tuned using oracle annotations and replay buffer samples. 
    }
    \label{fig:overview}
    \vspace{-8pt}
\end{figure*}

\section{Preliminaries} \label{sec:background}
We consider the standard RL setup~\cite{sutton2018reinforcement}, where an agent interacts with an environment in discrete time. At each timestep $t$, the agent observes a state $s_t$ and selects an action $a_t$ according to its policy $\pi$. The environment then provides a reward $r_t$ and transitions the agent to the next state $s_{t+1}$. The return $R_t = \sum_{k=0}^\infty \gamma^k r_{t+k}$ represents the discounted sum of future rewards starting at $t$, where $\gamma \in [0, 1)$ is the discount factor. The goal of RL is to maximize the expected return from each state $s_t$.

\noindent \textbf{Preference-based RL.} \label{sec:pref}
In the PbRL setup, an oracle provides feedback in the form of preferences between trajectory segments of the agent. The agent uses these preferences to construct an internal reward model $r_\theta$.
Formally, a trajectory segment $\sigma$ is defined as a sequence of observations and actions: $\sigma = \{(s_1, a_1), (s_2, a_2),...,(s_T, a_T)\}$. 
Given a pair of segments $(\sigma_0, \sigma_1)$, preferences are expressed as $y \in \{(0,1), (0.5, 0.5), (1,0)\}$, where $(1,0)$ indicates $\sigma_0 \succ \sigma_1$, $(0,1)$ indicates $\sigma_1 \succ \sigma_0$, and $(0.5,0.5)$ indicates a tie. Here, $\sigma_i \succ \sigma_j$ denotes that segment $i$ is preferred to segment $j$. The probability of one segment being preferred over another is modeled via the Bradley-Terry model~\cite{bradley1952rank}:
\begin{equation}
\label{eq:btmodel}
    P_{\theta}[\sigma_1 \succ \sigma_0] = \frac{\exp{\sum_{t}} r_{\theta}(s_t^1, a_t^1)}{\sum_{i\in\{0,1\}}\exp{\sum_{t}} r_{\theta}(s_t^i, a_t^i)} \ .
\end{equation}
To train the reward function $r_\theta$, we minimize the cross-entropy loss between the predicted preferences $P_{\theta}$ and the observed preferences $y$,
\begin{equation}\label{eq:btloss}
    \mathcal{L}_{\text{pref}} = - \mathbb{E}_{(\sigma_0,\sigma_1,y)\sim D} \big[y(0)P_\theta[\sigma_0 \succ\sigma_1] + y(1)P_\theta[\sigma_1\succ\sigma_0]\big] \ .
\end{equation}
The policy $\pi_\phi$ and the reward function $r_\theta$ are updated in an alternating fashion. First, the reward function is updated using the sampled preferences. Next, the policy is optimized to maximize the expected cumulative reward under the learned reward function using standard RL algorithms. 
\vspace{2pt}

\noindent \textbf{VLE as a reward model.} \label{sec:vlm}
VLE models comprise a language encoder $\mathcal{F}_L$ and an image encoder $\mathcal{F}_I$, which map text and image inputs into a shared latent space respectively. 
Using contrastive learning on image-caption pairs, often augmented with task-specific or dynamics-aware objectives~\cite{ma2023liv}, VLEs align textual and visual representations effectively. Given the image observation $o_t$ corresponding to a state $s_t$, and the language description of the task $l$, most works~\cite{ma2023liv,lidecisionnce} define the reward as:
\begin{equation}
\label{eq:cosine_sim}
    r^{vle}_t = \frac{\langle \mathcal{F}_{L}(l), \mathcal{F}_{I}(o_t) \rangle}{\Vert \mathcal{F}_{L}(l) \Vert \cdot \Vert \mathcal{F}_{I}(o_t) \Vert} \ .
\end{equation}
While this reward captures some aspects of task progress, it is often too coarse for fine-grained tasks such as manipulation. Although higher rewards tend to correspond to states closer to task completion, the signal is noisy and does not fully reflect nuanced progress~\cite{fu2024furl}. This motivates our approach: VLEs provide a scalable starting point for generating preference labels, which can then be refined with targeted oracle supervision. By improving VLE accuracy selectively, we reduce the need for expensive preference queries while maintaining high-quality reward signals.

\section{Method} \label{sec:method}
In this section, we present \algo, a framework designed to minimize the reliance on expensive oracle supervision in preference-based RL by leveraging VLEs. 
We first outline how VLEs are utilized to provide preference feedback for training a reward model (Sec.~\ref{sec:method_vlm_as_pref}). Then, we address the limitations of directly applying VLEs to new environments and propose a data-efficient adaptation approach that combines self-supervised learning with minimal oracle feedback to align the VLE with the environment's dynamics (Sec.~\ref{sec:method_vlm_adapt}). 
Finally, to ensure robust training in the presence of noisy machine-generated feedback, we introduce a filtering mechanism that identifies noisy VLE predictions and refines them with targeted oracle input - both to reduce annotation cost and to correct unreliable preference labels (Sec.~\ref{sec:method_sample_select}). Fig.~\ref{fig:overview} provides an overview of our approach.

\subsection{Scalable preference generation using VLE} \label{sec:method_vlm_as_pref}
To leverage VLEs for generating preference feedback, we begin by extracting image sequences corresponding to each trajectory segment. 
Specifically, for a given pair of segments $(\sigma_0, \sigma_1)$, we extract the image sequences $(O_0, O_1)$, where each sequence is defined as $O_i = \{o^i_0, o^i_1, o^i_2, \dots, o^i_T\}$ for $i \in \{0, 1\}$. 
Here, $o_t$ represents the image observation of the state $s_t$ at the $t$-th time step.

Using the language description of the task $l$, we compute the return for each segment as $R_i = \sum_{t=0}^T r^{vle}_t$, with $r^{vle}_t$ the reward at time step $t$ derived from the VLE via Eq. \ref{eq:cosine_sim}. 
Based on these returns, the preference label $\tilde{y}$ is assigned as: 
\begin{equation} \label{eq:vlm_pref} 
    \tilde{y} = \begin{cases} 
        (0, 1), & \text{if } R_1 > R_0, \\ 
        (1, 0), & \text{if } R_1 < R_0, \\ 
        (0.5, 0.5), & \text{otherwise}. 
    \end{cases} 
\end{equation}
These preferences are then used to train a reward model $r_\theta(s_t, a_t)$ by minimizing the preference loss in Eq.~\ref{eq:btloss}. 
The trained reward model can be integrated into a PbRL algorithm for policy optimization. 
In this work, we specifically leverage PEBBLE~\cite{lee2021pebble}, a PbRL framework that combines unsupervised pre-training with off-policy learning using Soft Actor-Critic~\cite{haarnoja2018soft}.

\begin{figure*}[t]
    \centering
    \includegraphics[width=\textwidth]{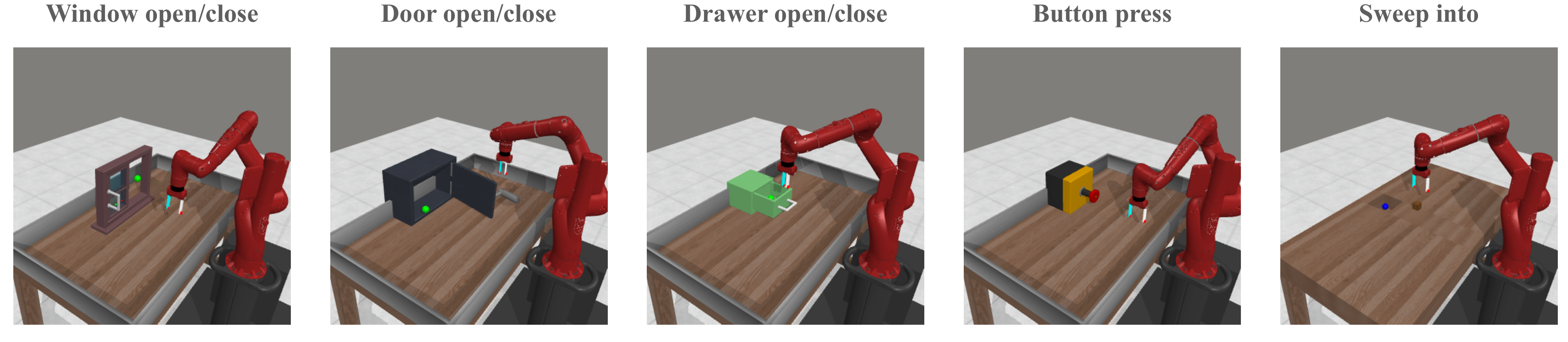}
    \caption{\textbf{Tasks.} 
    An illustration of the different manipulation tasks from Meta-World on which we evaluate our approach.
    }
    \label{fig:tasks}
    \vspace{-8pt}
\end{figure*}

\subsection{VLE adaptation to improve preference quality} \label{sec:method_vlm_adapt}
A key challenge in applying VLEs to downstream RL tasks is the domain gap between pretraining data and the target environment~\cite{raychaudhuri2021cross,fu2024furl}, which often leads to noisy feedback.
We address this with a data-efficient adaptation strategy that aligns the VLE to environment dynamics using minimal oracle feedback and self-supervised learning.

We freeze the VLE and introduce two learnable layers, $\mathcal{G}_L$ and $\mathcal{G}_I$, on top of the language and image embedding layers, respectively. 
These layers are fine-tuned to adapt the VLE. 
The layer $\mathcal{G}_L$ processes the language embedding $\mathcal{F}_L(l)$ of the task description $l$ and outputs an adapted text embedding, $\mathcal{G}_L\circ\mathcal{F}_L(l)$. 
Similarly, for each image observation $o_t$, the adapted image embedding is generated by $\mathcal{G}_I\circ\mathcal{F}_I(o_t)$. 
With these adapted embeddings, the reward for preference feedback is updated as: 
\begin{equation} \label{eq:cosine_sim_aligned} 
    r^{vle}_t = \frac{\langle \mathcal{G}_L\circ\mathcal{F}_L(l), \mathcal{G}_I\circ\mathcal{F}_I(o_t) \rangle}{\Vert \mathcal{G}_L\circ\mathcal{F}_L(l) \Vert \cdot \Vert \mathcal{G}_I\circ\mathcal{F}_I(o_t) \Vert}. 
\end{equation}
A small number of oracle-provided preferences are sampled to fine-tune the VLE using the preference loss in Eq.~\ref{eq:btloss}. 
The dense rewards for training are derived from the updated similarity measure (Eq.~\ref{eq:cosine_sim_aligned}). 
The methodology for selecting oracle feedback samples is detailed in Sec.~\ref{sec:method_sample_select}.

We also fine-tune the VLE using an unsupervised objective designed to align the VLE embeddings with environment dynamics. 
Given the current observation $o_t$, action $a_t$, and the next observation $o_{t+1}$, we train the VLE to predict the action which leads to the transition between observations: 
\begin{equation} \label{eq:inv_loss} 
    \Vert f\left(\mathcal{G}_I\circ\mathcal{F}_I(o_t), \mathcal{G}_I\circ\mathcal{F}_I(o_{t+1})\right) - a_t \Vert^2, 
\end{equation} 
where $f$ is a linear layer. 
This encourages the adapted embeddings to capture task-relevant dynamics, improving the precision of preference feedback.

\begin{figure*}[t]
    \centering
    \includegraphics[width=\textwidth]{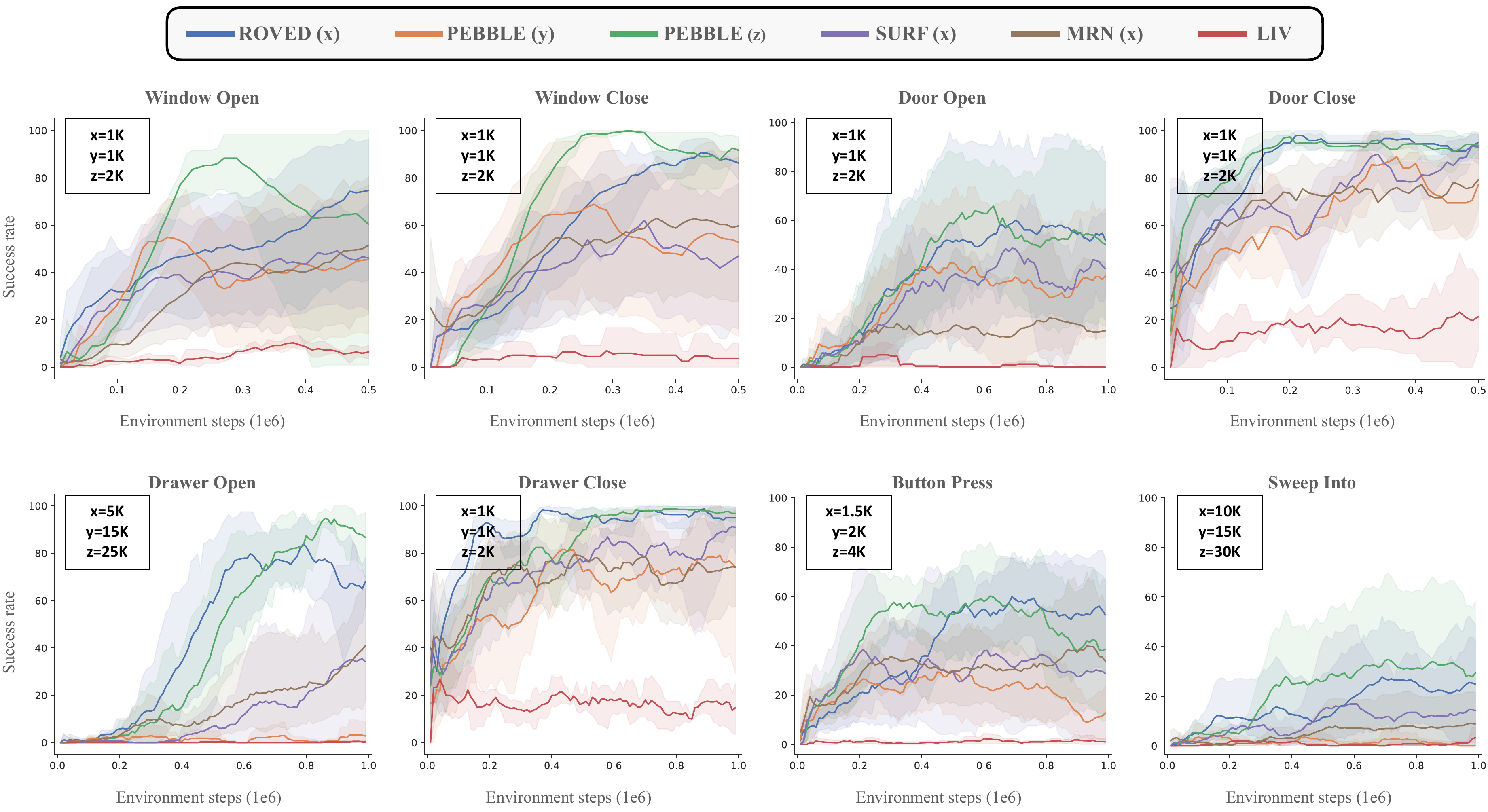}
    \caption{\textbf{Improved feedback efficiency.} 
    \algo~consistently outperforms all baselines with minimal oracle feedback, matching or exceeding PEBBLE’s performance while requiring 50\%-80\% fewer annotations. At equal preference counts, \algo~also outperforms MRN and SURF. Variables (x, y, z) denote the number of oracle preferences used.}
    \label{fig:exp_main}
    \vspace{-8pt}
\end{figure*}

\subsection{Noise mitigation and targeted oracle feedback} \label{sec:method_sample_select}
The VLE generated preferences, while scalable, are prone to noise and lack the reliability afforded by oracle-provided annotations. 
To ensure robust training, it is thus crucial to distinguish between accurate and noisy samples. 
This categorization not only improves the stability of the reward model training but also optimizes the use of oracle feedback by focusing on refining the noisy samples. 

\noindent \textbf{Identifying noisy samples.} 
Our approach builds on findings from noisy training~\cite{cheng2024rime}, which show that deep networks fit clean samples before memorizing noisy labels. We prioritize samples with lower training losses as clean, while treating high-loss samples as potential candidates for oracle review. 

Specifically, consider the preference loss defined in Eq.~\ref{eq:btloss} for training the reward model $r_\theta$. Assuming the loss for clean samples is bounded by $\rho$, Cheng \textit{et al.}~\cite{cheng2024rime} show that the KL divergence between the predicted preference distribution $P_\theta$ and label $\tilde{y}$ for a noisy sample $(\sigma_0, \sigma_1)$ is lower bounded as:
\begin{equation}\label{eq:KL_div} 
    D_{KL}(\tilde{y}\Vert P_\theta) \geq -\ln{\rho} + \frac{\rho}{2} + \mathcal{O}(\rho^2) \ . 
\end{equation}
To filter out unreliable samples based on Eq.~\ref{eq:KL_div}, we take a lower bound on the KL divergence, $\tau_{base} = -\ln{\rho} + \alpha \rho$, where $\rho$ is the maximum loss calculated on the filtered samples from the latest update, and $\alpha \in (0, 0.5]$ is a hyperparameter. 
To ensure tolerance for clean samples under distribution shifts during policy learning, we introduce an additional uncertainty term in the lower bound, $\tau_{\text{unc}} = \beta_t \cdot s_{KL}$, where $s_{KL}$ is the standard deviation of KL divergence across encountered samples, and $\beta_t$ decays linearly over time. Specifically, $\beta_t = \max(\beta_{min}, \beta_{max} - kt)$, where $\beta_{\min}$, $\beta_{\max}$, and $k$ are hyperparameters controlling the decay. The final dynamic lower bound is given by $\tau_{lower} = \tau_{base} + \tau_{unc}$. 

This adaptive thresholding enables greater tolerance for noisy samples early in training, while becoming more conservative as learning progresses.

\noindent \textbf{Handling noisy samples.} Samples with a KL divergence below $\tau_{lower}$ are considered clean and are incorporated into the reward model training: 
\begin{equation} 
\label{D_lower}
    \mathcal{D}_{\tau_{l}} = \{(\sigma^0, \sigma^1, \tilde{y}) : D_{KL}(\tilde{y} \Vert P_\theta(\sigma_0, \sigma_1)) < \tau_{lower}\} \ . 
\end{equation}
Conversely, samples with a KL divergence exceeding a higher threshold $\tau_{upper}$ are presumed to contain noisy labels with high certainty. To preserve their utility, we relabel these samples by flipping their predicted labels and include them in a separate dataset:
\begin{equation}
\label{D_upper}
    \mathcal{D}_{\tau_u} = \{(\sigma_0, \sigma_1, \mathbf{1}-\tilde{y}) : D_{KL}(\tilde{y} \Vert P_\theta(\sigma_0, \sigma_1)) > \tau_{upper}\} \ .
\end{equation}
The remaining samples, with KL divergence between $\tau_{lower}$ and $\tau_{upper}$, are deemed \emph{uncertain}, and we sample from them based on the available oracle annotation budget. 
These samples are particularly valuable, as both the VLE and reward model struggle to assign reliable labels. 
By focusing annotation on this subset, $\mathcal{D}_{o}$, we ensure that annotations address the most uncertain cases. 

\noindent \textbf{Training with selective feedback.} 
The reward model is trained on $N_{\text{vle}} = |\mathcal{D}_{\tau_l}| + |\mathcal{D}_{\tau_u}|$ machine-labeled samples, supplemented by $N_{\text{oracle}}=|\mathcal{D}_o|$ oracle-labeled samples from uncertain cases. 
The $N_{\text{oracle}}$ samples are used to update the VLE. 
This targeted feedback mechanism improves the reward model and the VLE while significantly reducing the overall annotation burden. 

\subsection{Overall algorithm} \label{sec:method_overall}
\algo~proceeds by initializing the policy $\pi_\phi$, reward function $r_\theta$, and additional layers $\mathcal{G}$ on top of the VLE randomly. 
Given a task description $l$, the method iteratively follows a structured cycle. 
First, the policy $\pi_\phi$ is updated using the reward function $r_\theta$, interacting with the environment and storing observed trajectories in a buffer. From this buffer, trajectory segment pairs are sampled and assigned preferences using the VLE using Eq.~\ref{eq:vlm_pref} and Eq.~\ref{eq:cosine_sim_aligned}. These labeled pairs are then categorized into clean and noisy samples based on the filtering strategy outlined in Sec.~\ref{sec:method_sample_select}. 
A subset of the noisy samples is sent for oracle annotation, subject to a predefined budget. 
The reward model is updated using the preference-labeled pairs ( VLE/oracle annotated) through Eq.~\ref{eq:btloss}, while the VLE is fine-tuned using the oracle-annotated samples, following the adaptation strategy outlined in Sec.~\ref{sec:method_vlm_adapt}. 

\section{Experiments} \label{sec:experiments}







\subsection{Experimental setup} \label{sec:exp_setup}
\noindent \textbf{Tasks.} 
We evaluate \algo~on eight diverse manipulation tasks from Meta-World~\cite{yu2020meta}: \emph{door-open}, \emph{door-close}, \emph{drawer-open}, \emph{drawer-close}, \emph{window-open}, \emph{window-close}, \emph{button-press}, and \emph{sweep-into}. The tasks are visualized in Fig.~\ref{fig:tasks} 

\noindent \textbf{Implementation.} 
We use Soft Actor-Critic (SAC)~\cite{haarnoja2018soft} with state-based observations. Actor and critic models are MLPs (3 hidden layers, 256 units) trained with Adam using a learning rate (LR) of $1\mathrm{e}{-4}$; the critic $\tau$ is set to the default value of $5\mathrm{e}{-3}$.  The reward model is an ensemble of 3 MLPs (3 hidden layers, 256 units, Leaky ReLU; tanh output), trained with Adam (LR $3\mathrm{e}{-4}$, batch size 128) for 200 steps per iteration. The reward model is trained on 30K preference samples in total, combining the allocated oracle preference budget with the remaining preferences generated by the VLE.

We collect 1K random interactions, followed by 5K unsupervised interactions using state-entropy maximization. Policy, reward model, and VLE training begin after these 6K steps. For initialization, 250 oracle feedback samples are used to train both the reward model and the VLE. Subsequently, at every 3K steps, $N=128$ preference pairs are sampled using segments of length 50 from the SAC replay buffer, following PEBBLE~\cite{lee2021pebble}. VLE labels for these pairs are added to the preference buffer $\mathcal{B}$. From $\mathcal{B}$, we sample $\mathcal{D}_{\tau_{l}}$ and $\mathcal{D}_{\tau_{u}}$ as described in Sec.~\ref{sec:method_sample_select}. From the remaining uncertain pairs, up to $0.25N$ are selected for oracle feedback. This procedure is repeated every 3K steps until either the total preference budget or the oracle preference budget is exhausted.

To determine the oracle feedback budget, we first ran PEBBLE with varying budgets (1K–30K) across tasks, and identified the minimum budget to achieve a reasonable success rate ($>50\%$) in each task. Although larger budgets generally improve performance, our goal is to show that incorporating VLEs can substantially reduce the amount of oracle feedback needed to reach comparable results. We set task-specific oracle budgets based on task complexity.

We experiment with both LIV~\cite{ma2023liv} and DecisionNCE~\cite{lidecisionnce} as VLEs, and implement \algo~with LIV by default unless specified. The trainable layers, $\mathcal{G}_L$ and $\mathcal{G}_I$, are 2-layer MLPs (256, 64, ReLU), while the inverse dynamics head $f$ is a 128–64–4 MLP with ReLU, with an LR of $3\mathrm{e}{-4}$. For label selection, we adopt RIME~\cite{cheng2024rime} hyperparameters: $\alpha=0.5$, $\beta_{min}=1$, $\beta_{max}=3$, $k=1/300$, and $\tau_{upper}=3\ln(10)$.  

\noindent \textbf{VLM oracles.} 
We also evaluate using a VLM-based oracle, following RL-VLM-F~\cite{wang2024} with Gemini-Pro-1.5~\cite{team2024gemini}. Due to API costs\footnote{\url{https://ai.google.dev/gemini-api/docs/pricing\#gemini-1.5-pro}}, results are reported on a subset of tasks. We adopt the exact RL-VLM-F setup (segment length 1, two-step prompting, 20K budget), and additionally evaluate with a reduced budget of 10K. Note that RL-VLM-F requires more queries than PEBBLE as it operates on single-step segments. 

\noindent \textbf{Baselines.} 
We compare \algo~against several baselines. \textit{PEBBLE}~\cite{lee2021pebble} serves as a basic reference, learning a reward function from human preference feedback and optimizing policies using SAC. SURF~\cite{park2022surf} and MRN~\cite{liu2022meta} improve preference learning efficiency via semi-supervised augmentation and policy-informed reward updates, respectively. For pure VLE-based approaches, we include \textit{LIV}~\cite{ma2023liv} and \textit{DecisionNCE}~\cite{lidecisionnce}. Finally, \textit{RL-VLM-F}~\cite{wang2024} is evaluated as a VLM-based oracle. All results are averaged across five random seeds.  

\begin{figure}[t]
\centering
\includegraphics[width=0.49\textwidth]{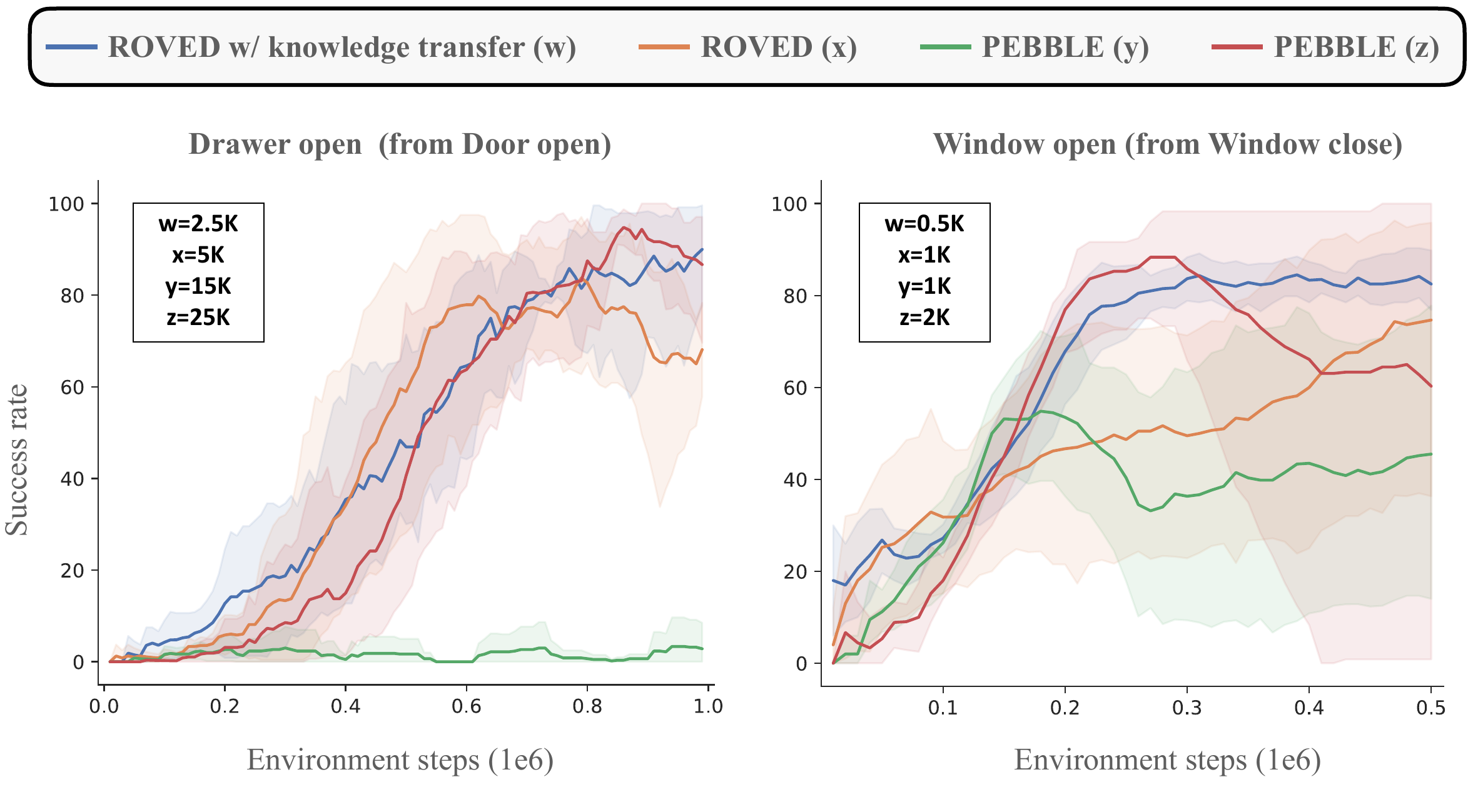}
\caption{\textbf{Knowledge transfer across tasks.} 
With knowledge transfer, \algo~matches or surpasses PEBBLE while reducing annotation requirements by 75–90\%. This demonstrates effective transfer in both \emph{same task, different object} (left) and \emph{same object, different task} (right) settings. Variables (w, x, y, z) denote the number of preferences used.}
\label{fig:transfer_results}
\vspace{-8pt}
\end{figure}

\begin{figure}[t]
\centering
\includegraphics[width=0.49\textwidth]{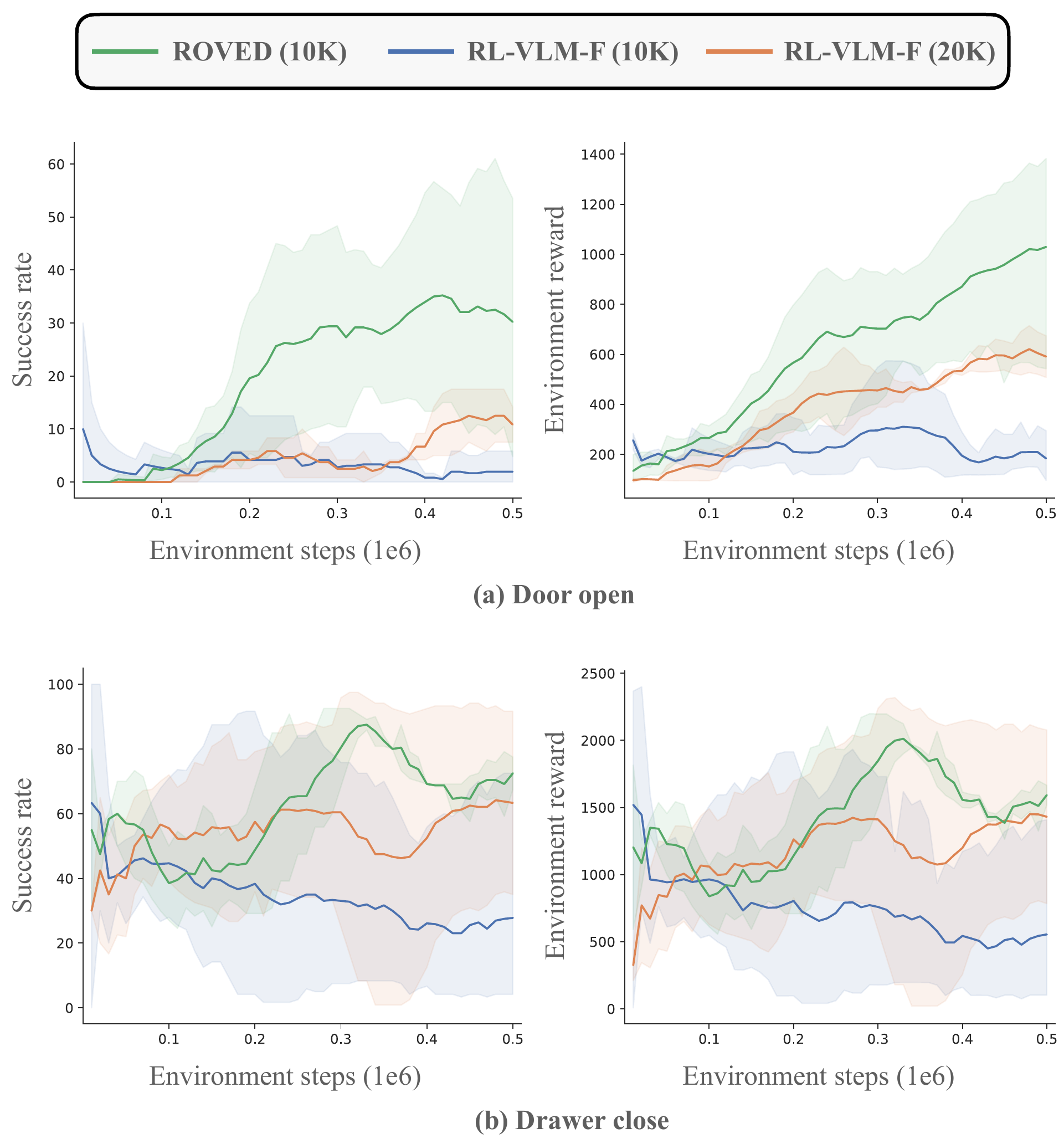}
\caption{\textbf{Experiments with VLM oracles.} 
\algo~achieves comparable or better performance than RL-VLM-F while using $50\%$ fewer oracle preferences (denoted by the number in brackets), demonstrating its ability to generalize across different PbRL algorithms. Experiments are reported on a subset of environments due to the high API cost of VLMs.} 
\label{fig:gemini}
\vspace{-8pt}
\end{figure}

\subsection{Main results} \label{sec:main_results}
\noindent \textbf{Does \algo~improve oracle feedback efficiency?} 
We evaluate whether \algo~can achieve high task performance with substantially less oracle feedback. Fig.~\ref{fig:exp_main} shows the learning curves comparing task success rates. \algo~is evaluated using $x$ oracle feedback samples, while PEBBLE, the baseline preference-based method, is assessed with $y$ (similar or higher than $x$) and $z$ (significantly higher) samples to highlight the impact of feedback efficiency. Efficiency-focused PbRL methods, SURF and MRN, are also evaluated using $x$ oracle samples to examine the gains by \algo.

Across most tasks, \algo~matches PEBBLE's performance while requiring only 50\% fewer oracle queries. For more challenging tasks (\textit{e.g.}, button-press, sweep-into, and drawer-open), the reduction increases to 63.5\%, 66.6\%, and 80\%, respectively. Standalone VLE-based reward models (LIV and DecisionNCE) fail to achieve meaningful success due to noisy rewards when applied without supervision (see Sec.~\ref{sec:vlm}), highlighting the need for minimal oracle feedback. While SURF and MRN show moderate gains over PEBBLE at reduced oracle budgets, they still lag behind \algo~across all tasks, emphasizing the effectiveness of our approach.
\begin{figure}[t]
    \centering
    \includegraphics[width=0.49\textwidth]{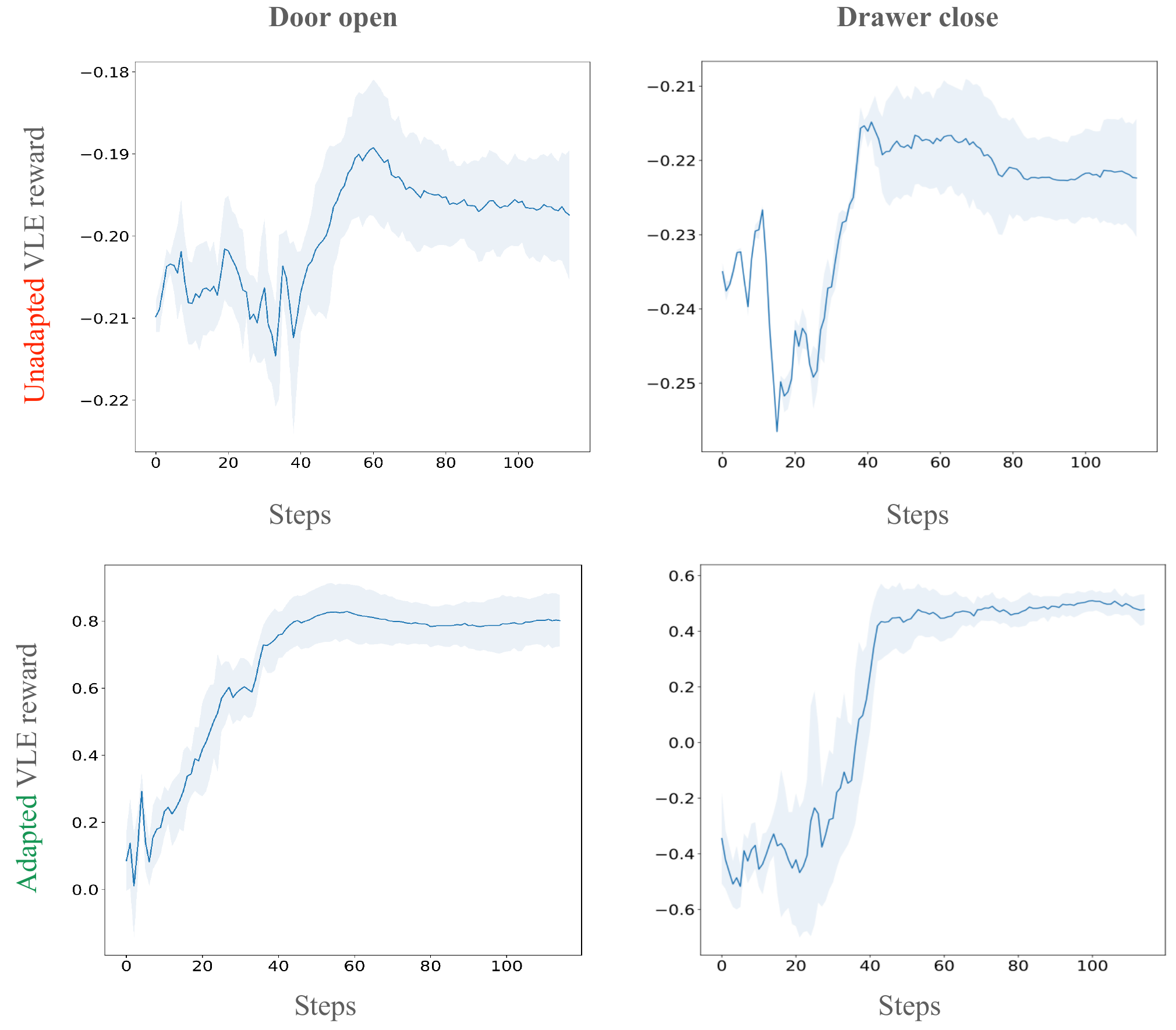}
    \caption{\textbf{Impact of VLE adaptation. }
    VLE reward predictions on the door-open and drawer-close tasks before (top row) and after (bottom row) adaptation, averaged over the same 5 expert trajectories. After adaptation, the VLE aligns more closely with true task progress.}
    \label{fig:adapted_vlm}
    \vspace{-8pt}
\end{figure}

\noindent \textbf{Can \algo~transfer knowledge across tasks?} 
A key objective of \algo~is to refine the VLE with oracle feedback to reduce inherent noise. We test whether an adapted VLE can generalize to related tasks with minimal additional supervision. Two types of transfer are considered: (1) \emph{same task, different object}: ``door-open'' $\rightarrow$ ``drawer-open''; (2) \emph{same object, different task}: ``window-close'' $\rightarrow$ ``window-open''. In both cases, the VLE for the target task is initialized with weights from the source task, while the rest of the algorithm remains unchanged. Fig.~\ref{fig:transfer_results} shows that \algo~with knowledge transfer matches PEBBLE trained with 25K queries using as few as 2.5K, a $90\%$ reduction in oracle queries. Even in the worst case, we observe at least 75\% query reduction. This transferability enables cumulative annotation savings across tasks, a feature absent in existing methods.

\noindent \textbf{Is \algo~effective with large vision-language models as oracles?} 
We also evaluate \algo~in settings where large VLMs act as automated oracles, following the RL-VLM-F setup~\cite{wang2024} with Gemini-Pro-1.5~\cite{team2024gemini}. As shown in Fig.~\ref{fig:gemini}, \algo~reduces the number of oracle queries by 50\% without compromising policy performance and improves cumulative episode rewards. Although VLMs provide accurate preference labels, they are slow ($\sim$5s per query) and incur high API costs. \algo~leverages lightweight VLEs to automatically generate most preference labels, reserving expensive VLM queries for uncertain cases. This approach produces feedback in under 0.009s—over three orders of magnitude faster—while cutting reliance on costly oracle queries by at least 50\%, demonstrating that VLEs are essential for efficient PbRL.

\begin{figure}[t]
    \centering
    \includegraphics[width=0.45\textwidth]{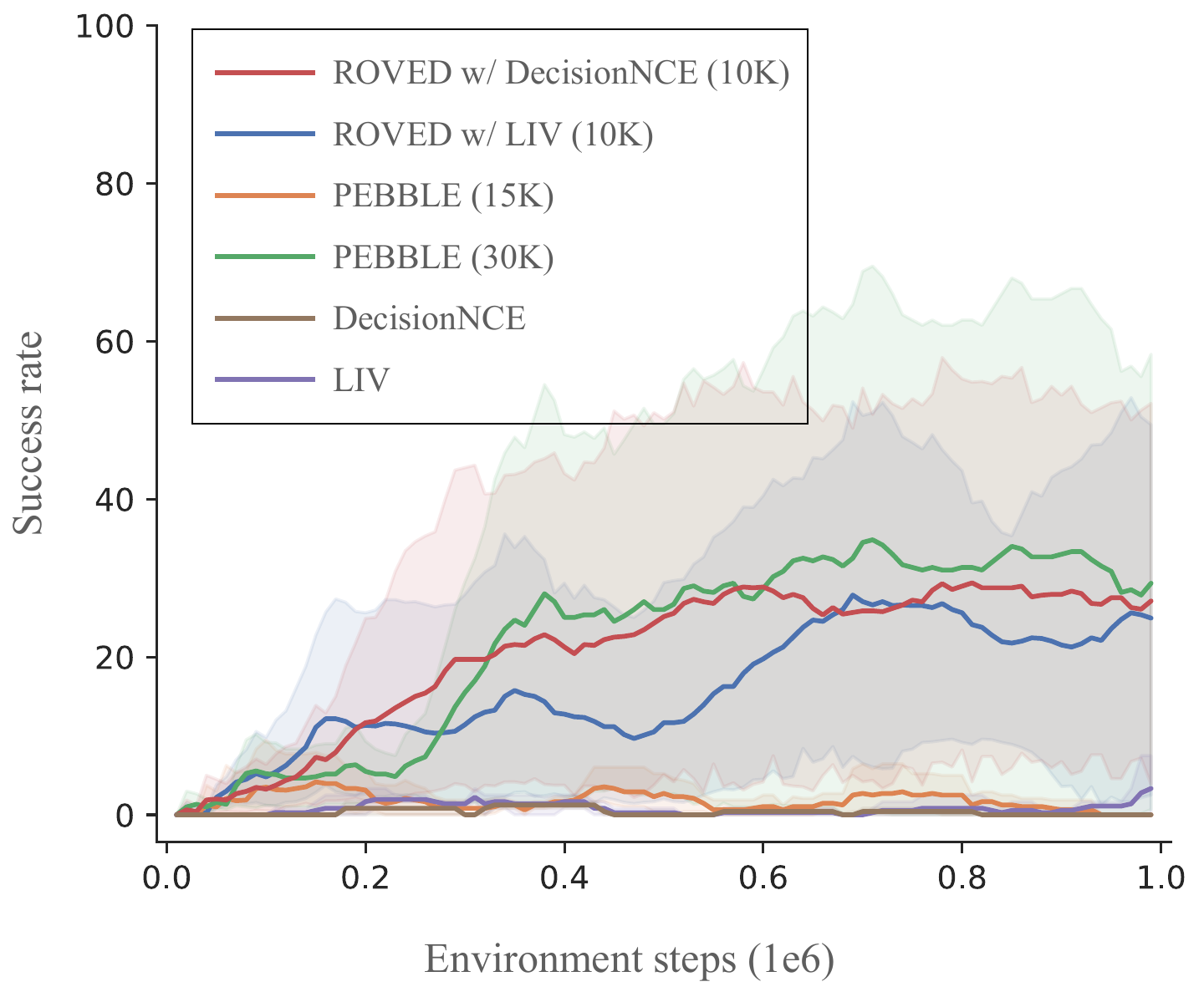}
    \caption{\textbf{Robustness to choice of VLE. } 
    \algo~maintains consistent performance (on the sweep-into task) across different VLE backbones, including LIV and DecisionNCE, highlighting robustness to the choice of embedding model.}
    \label{fig:exp_vle}
\end{figure}

\begin{figure}[t]
    \centering
    \includegraphics[width=0.49\textwidth]{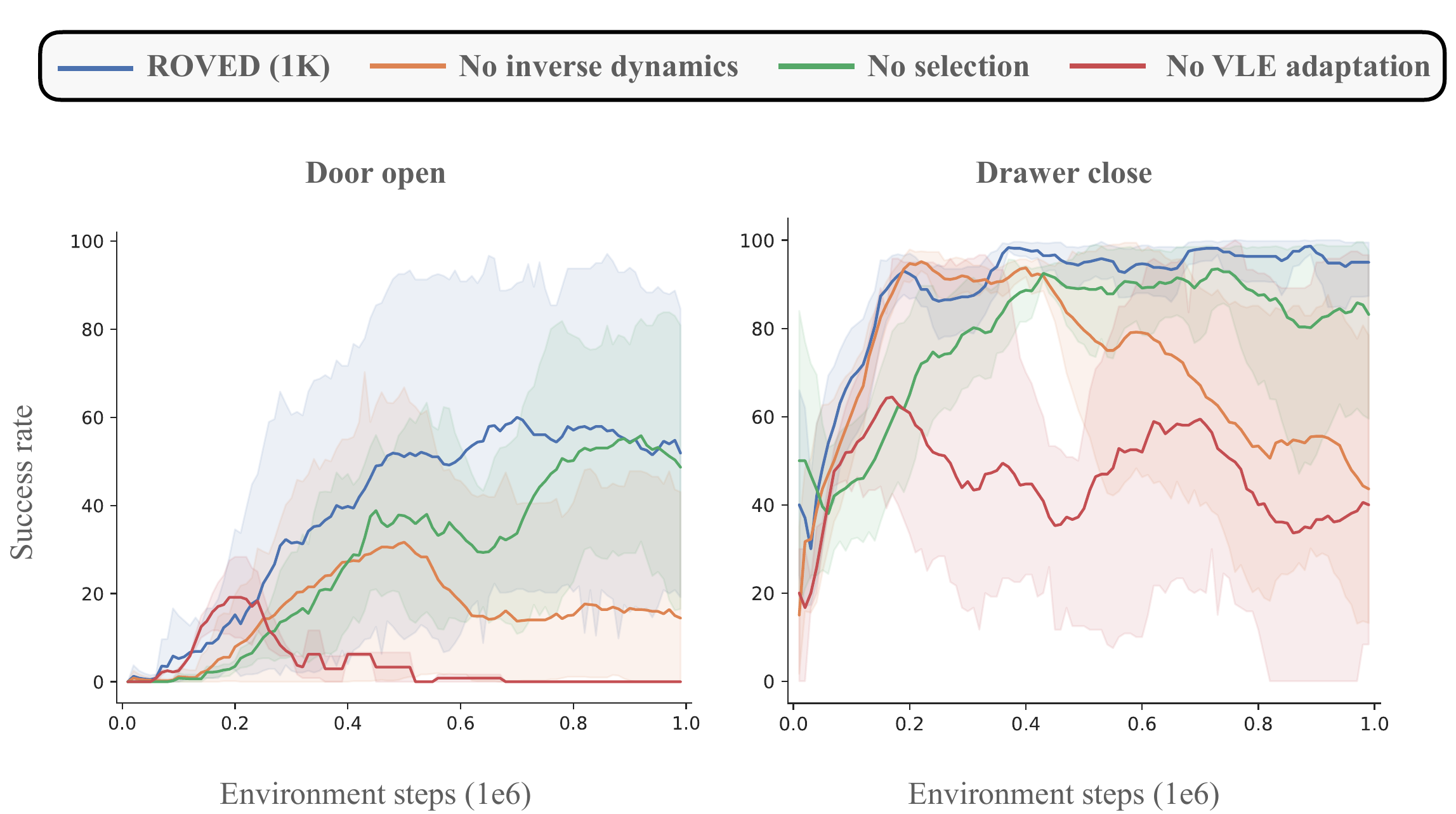}
    \caption{\textbf{Ablation study.} Success rates when (i) removing the inverse dynamics loss from VLE adaptation, (ii) replacing selective feedback with random sampling or (iii) removing VLE adaptation. All ablations show clear performance degradation, underscoring the importance of each component.}
    \label{fig:ablation}
    \vspace{-8pt}
\end{figure}

\subsection{Ablation study} \label{sec:ablation}

\noindent \textbf{Impact of VLE adaptation.} 
To understand how oracle feedback improves the VLE, we analyze reward outputs before and after adaptation. Fig.~\ref{fig:adapted_vlm} shows VLE reward predictions on five expert trajectories for the door-open and drawer-close tasks. Ideally, rewards should increase along the trajectory, reflecting progress toward task completion. As shown, the adapted VLE better aligns with ground-truth task progress, producing higher reward values toward the end of successful trajectories without abrupt drops. This reduces reliance on expensive oracle feedback and enables the VLE to serve as a more effective and generalizable prior. Consequently, fine-tuned VLEs transfer across tasks more robustly, compounding annotation savings over time (see Sec.~\ref{sec:main_results}). To further analyze the impact of this adaptation, we trained only the reward model using feedback from the VLE and the oracle, while keeping the VLE fixed (Sec.~\ref{sec:method_vlm_adapt}). As shown in Fig.~\ref{fig:ablation}, although the success rate initially increases, it later drops, since the VLE does not incorporate oracle feedback, leading to increasingly misaligned reward signals.

\noindent \textbf{Robustness to choice of VLE.} 
Our main experiments use LIV~\cite{ma2023liv} as the default VLE, but we also evaluate \algo~with DecisionNCE~\cite{lidecisionnce} to test sensitivity to the choice of VLE. As shown in Fig.~\ref{fig:exp_vle} on the sweep-into task, \algo~successfully leverages stronger pre-trained VLEs: DecisionNCE slightly improves performance over LIV. Importantly, DecisionNCE alone fails to solve these tasks without \algo’s uncertainty-aware hybrid components, validating the necessity of selective adaptation and oracle-guided feedback. This shows that \algo’s benefits generalize across VLE backbones and makes the approach adaptable to stronger or more specialized VLEs in future work.

\noindent \textbf{Impact of inverse dynamics loss.}  
To evaluate the efficacy of the self-supervised loss, we trained the VLE using only the preference loss in Eq.~\ref{eq:btloss}. As shown in Fig.~\ref{fig:ablation}, the success rate initially increases but later degrades. In the early stages, when oracle feedback is available, the VLE performs well by leveraging reliable supervision. However, without the inverse dynamics loss, its performance declines as the policy evolves. This occurs because, while the VLE learns from oracle preferences, these preferences are derived from samples of a suboptimal policy. As the policy improves and the data distribution shifts, the VLE struggles to generalize due to limited understanding of environment dynamics. Incorporating the inverse dynamics loss allows the VLE to adapt to distribution shifts, maintaining stable performance throughout.

\noindent \textbf{Impact of selective feedback.} 
To evaluate the efficacy of the selection strategy, we evaluate a variant of \algo~where oracle feedback is collected on randomly selected pairs from the replay buffer instead of using the noise mitigation and selection strategy in Sec.~\ref{sec:method_sample_select}. As shown in Fig.~\ref{fig:ablation}, removing this selection achieves reasonable performance and can even surpass PEBBLE with the same amount of feedback, but remains inferior to the full method. Random selection allows noisy labels to influence reward model training and introduces redundancy in oracle feedback, as uncertain segments for the VLE may not be prioritized. This shows the need of selective feedback in maximizing oracle efficiency.

\section{Conclusion}
In this work, we introduced \algo, a hybrid framework for PbRL that combines the scalability of VLEs with the precision of selective oracle supervision. By leveraging an uncertainty-aware sample selection strategy and a parameter-efficient VLE adaptation mechanism, we reduce reliance on costly human or VLM feedback. Across several manipulation tasks, we achieve comparable or superior policy performance while cutting annotation costs by 50–80\%. Moreover, our fine-tuned VLE transfers effectively across related tasks, enabling cumulative annotation savings of 75–90\%. These results highlight \algo~as a practical step toward more efficient and scalable PbRL.

\noindent \textbf{Acknowledgement.} 
This work was partially supported by NSF grants under Award No. 2326309 and 2312395, and the UC Multi-campus Research Programs Initiative.

\bibliographystyle{IEEEtran}
\bibliography{ref}

@article{ho2016generative,
  title={Generative adversarial imitation learning},
  author={Ho, Jonathan and Ermon, Stefano},
  journal={NeurIPS},
  year={2016}
}

@article{rocamonde2023vision,
  title={Vision-language models are zero-shot reward models for reinforcement learning},
  author={Rocamonde, Juan and Montesinos, Victoriano and Nava, Elvis and Perez, Ethan and Lindner, David},
  journal={arXiv preprint arXiv:2310.12921},
  year={2023}
}

@inproceedings{ma2023liv,
  title={Liv: Language-image representations and rewards for robotic control},
  author={Ma, Yecheng Jason and Kumar, Vikash and Zhang, Amy and Bastani, Osbert and Jayaraman, Dinesh},
  booktitle={ICML},
  year={2023},
}

@article{sontakke2024roboclip,
  title={Roboclip: One demonstration is enough to learn robot policies},
  author={Sontakke, Sumedh and Zhang, Jesse and Arnold, S{\'e}b and Pertsch, Karl and B{\i}y{\i}k, Erdem and Sadigh, Dorsa and Finn, Chelsea and Itti, Laurent},
  journal={NeurIPS},
  year={2024}
}

@inproceedings{radford2021learning,
  title={Learning transferable visual models from natural language supervision},
  author={Radford, Alec and Kim, Jong Wook and Hallacy, Chris and Ramesh, Aditya and Goh, Gabriel and Agarwal, Sandhini and Sastry, Girish and Askell, Amanda and Mishkin, Pamela and Clark, Jack and others},
  booktitle={ICML},
  year={2021},
}

@article{fu2024furl,
  title={FuRL: Visual-Language Models as Fuzzy Rewards for Reinforcement Learning},
  author={Fu, Yuwei and Zhang, Haichao and Wu, Di and Xu, Wei and Boulet, Benoit},
  journal={arXiv preprint arXiv:2406.00645},
  year={2024}
}

@inproceedings{hejna2023few,
  title={Few-shot preference learning for human-in-the-loop rl},
  author={Hejna III, Donald Joseph and Sadigh, Dorsa},
  booktitle={CoRL},
  year={2023},
}

@article{smith2019avid,
  title={Avid: Learning multi-stage tasks via pixel-level translation of human videos},
  author={Smith, Laura and Dhawan, Nikita and Zhang, Marvin and Abbeel, Pieter and Levine, Sergey},
  journal={arXiv preprint arXiv:1912.04443},
  year={2019}
}

@article{akgun2012keyframe,
  title={Keyframe-based learning from demonstration: Method and evaluation},
  author={Akgun, Baris and Cakmak, Maya and Jiang, Karl and Thomaz, Andrea L},
  journal={International Journal of Social Robotics},
  volume={4},
  pages={343--355},
  year={2012},
  publisher={Springer}
}

@article{song2022learning,
  title={Learning from noisy labels with deep neural networks: A survey},
  author={Song, Hwanjun and Kim, Minseok and Park, Dongmin and Shin, Yooju and Lee, Jae-Gil},
  journal={IEEE transactions on neural networks and learning systems},
  volume={34},
  number={11},
  pages={8135--8153},
  year={2022},
  publisher={IEEE}
}

@article{cheng2024rime,
  title={RIME: Robust Preference-based Reinforcement Learning with Noisy Preferences},
  author={Cheng, Jie and Xiong, Gang and Dai, Xingyuan and Miao, Qinghai and Lv, Yisheng and Wang, Fei-Yue},
  journal={arXiv preprint arXiv:2402.17257},
  year={2024}
}

@book{sutton2018reinforcement,
  title={Reinforcement learning: An introduction},
  author={Sutton, Richard S and Barto, Andrew G},
  year={2018},
  publisher={MIT press}
}

@article{chiang2019learning,
  title={Learning navigation behaviors end-to-end with autorl},
  author={Chiang, Hao-Tien Lewis and Faust, Aleksandra and Fiser, Marek and Francis, Anthony},
  journal={IEEE RA-L},
  volume={4},
  number={2},
  pages={2007--2014},
  year={2019},
  publisher={IEEE}
}

@article{ma2023eureka,
  title={Eureka: Human-level reward design via coding large language models},
  author={Ma, Yecheng Jason and Liang, William and Wang, Guanzhi and Huang, De-An and Bastani, Osbert and Jayaraman, Dinesh and Zhu, Yuke and Fan, Linxi and Anandkumar, Anima},
  journal={arXiv preprint arXiv:2310.12931},
  year={2023}
}

@article{rafailov2024direct,
  title={Direct preference optimization: Your language model is secretly a reward model},
  author={Rafailov, Rafael and Sharma, Archit and Mitchell, Eric and Manning, Christopher D and Ermon, Stefano and Finn, Chelsea},
  journal={NeurIPS},
  year={2024}
}

@article{lee2021pebble,
  title={Pebble: Feedback-efficient interactive reinforcement learning via relabeling experience and unsupervised pre-training},
  author={Lee, Kimin and Smith, Laura and Abbeel, Pieter},
  journal={arXiv preprint arXiv:2106.05091},
  year={2021}
}

@article{christiano2017deep,
  title={Deep reinforcement learning from human preferences},
  author={Christiano, Paul F and Leike, Jan and Brown, Tom and Martic, Miljan and Legg, Shane and Amodei, Dario},
  journal={NeurIPS},
  year={2017}
}

@article{park2022surf,
  title={SURF: Semi-supervised reward learning with data augmentation for feedback-efficient preference-based reinforcement learning},
  author={Park, Jongjin and Seo, Younggyo and Shin, Jinwoo and Lee, Honglak and Abbeel, Pieter and Lee, Kimin},
  journal={arXiv preprint arXiv:2203.10050},
  year={2022}
}

@article{liu2022meta,
  title={Meta-reward-net: Implicitly differentiable reward learning for preference-based reinforcement learning},
  author={Liu, Runze and Bai, Fengshuo and Du, Yali and Yang, Yaodong},
  journal={NeurIPS},
  volume={35},
  year={2022}
}

@inproceedings{lukasik2020does,
  title={Does label smoothing mitigate label noise?},
  author={Lukasik, Michal and Bhojanapalli, Srinadh and Menon, Aditya and Kumar, Sanjiv},
  booktitle={ICML},
  year={2020},
}

@article{zhang2018generalized,
  title={Generalized cross entropy loss for training deep neural networks with noisy labels},
  author={Zhang, Zhilu and Sabuncu, Mert},
  journal={NeurIPS},
  year={2018}
}

@inproceedings{wang2021denoising,
  title={Denoising implicit feedback for recommendation},
  author={Wang, Wenjie and Feng, Fuli and He, Xiangnan and Nie, Liqiang and Chua, Tat-Seng},
  booktitle={Proceedings of the 14th ACM international conference on web search and data mining},
  pages={373--381},
  year={2021}
}

@article{wang2022self,
  title={Self-instruct: Aligning language models with self-generated instructions},
  author={Wang, Yizhong and Kordi, Yeganeh and Mishra, Swaroop and Liu, Alisa and Smith, Noah A and Khashabi, Daniel and Hajishirzi, Hannaneh},
  journal={arXiv preprint arXiv:2212.10560},
  year={2022}
}

@article{bradley1952rank,
  title={Rank analysis of incomplete block designs: I. The method of paired comparisons},
  author={Bradley, Ralph Allan and Terry, Milton E},
  journal={Biometrika},
  volume={39},
  number={3/4},
  pages={324--345},
  year={1952},
  publisher={JSTOR}
}

@inproceedings{haarnoja2018soft,
  title={Soft actor-critic: Off-policy maximum entropy deep reinforcement learning with a stochastic actor},
  author={Haarnoja, Tuomas and Zhou, Aurick and Abbeel, Pieter and Levine, Sergey},
  booktitle={ICML},
  year={2018},
}

@inproceedings{raychaudhuri2021cross,
  title={Cross-domain imitation from observations},
  author={Raychaudhuri, Dripta S and Paul, Sujoy and Vanbaar, Jeroen and Roy-Chowdhury, Amit K},
  booktitle={ICML},
  year={2021},
}

@inproceedings{yu2020meta,
  title={Meta-world: A benchmark and evaluation for multi-task and meta reinforcement learning},
  author={Yu, Tianhe and Quillen, Deirdre and He, Zhanpeng and Julian, Ryan and Hausman, Karol and Finn, Chelsea and Levine, Sergey},
  booktitle={CoRL},
  pages={1094--1100},
  year={2020}
}

@InProceedings{wang2024,
          title = 	 {RL-VLM-F: Reinforcement Learning from Vision Language Foundation Model Feedback},
          author =       {Wang, Yufei and Sun, Zhanyi and Zhang, Jesse and Xian, Zhou and Biyik, Erdem and Held, David and Erickson, Zackory},
          booktitle = 	 {ICML},
          year = 	 {2024}
        }

@article{wang2025prefclm,
  title={Prefclm: Enhancing preference-based reinforcement learning with crowdsourced large language models},
  author={Wang, Ruiqi and Zhao, Dezhong and Yuan, Ziqin and Obi, Ike and Min, Byung-Cheol},
  journal={IEEE RA-L},
  year={2025},
  publisher={IEEE}
}

@article{li2024survey,
  title={A survey on deep active learning: Recent advances and new frontiers},
  author={Li, Dongyuan and Wang, Zhen and Chen, Yankai and Jiang, Renhe and Ding, Weiping and Okumura, Manabu},
  journal={Transactions on Neural Networks and Learning Systems},
  year={2024}
}

@article{team2024gemini,
  title={Gemini 1.5: Unlocking multimodal understanding across millions of tokens of context},
  author={Team, Gemini and Georgiev, Petko and Lei, Ving Ian and Burnell, Ryan and Bai, Libin and Gulati, Anmol and Tanzer, Garrett and Vincent, Damien and Pan, Zhufeng and Wang, Shibo and others},
  journal={arXiv preprint arXiv:2403.05530},
  year={2024}
}

@inproceedings{lidecisionnce,
  title={DecisionNCE: Embodied Multimodal Representations via Implicit Preference Learning},
  author={Li, Jianxiong and Zheng, Jinliang and Zheng, Yinan and Mao, Liyuan and Hu, Xiao and Cheng, Sijie and Niu, Haoyi and Liu, Jihao and Liu, Yu and Liu, Jingjing and others},
  booktitle={ICML},
  year={2024}
}

@article{
ghosh2025robust,
title={Robust Offline Imitation Learning from Diverse Auxiliary Data},
author={Udita Ghosh and Dripta S. Raychaudhuri and Jiachen Li and Konstantinos Karydis and Amit Roy-Chowdhury},
journal={Transactions on Machine Learning Research},
issn={2835-8856},
year={2025},
}

\end{document}